\documentclass{article}

\usepackage[english]{babel}

\usepackage[letterpaper,top=2cm,bottom=2cm,left=3cm,right=3cm,marginparwidth=1.75cm]{geometry}

\usepackage{amsmath}
\usepackage{amsfonts} 

\usepackage[group-separator={,}]{siunitx} 
\usepackage{dsfont} 
\usepackage{bbding}

\usepackage{graphicx}
\graphicspath{ {./figs/} }

\usepackage[table]{xcolor} 
\usepackage{tabularx}
\usepackage{multirow} 
\usepackage{adjustbox} 

\usepackage{algorithm}
\usepackage{algorithmic}

\usepackage[colorlinks=true, linkcolor=blue, urlcolor=blue, citecolor=cyan, bookmarks=true, bookmarksopen,
bookmarksopenlevel=3,
pagebackref=true]{hyperref}
\usepackage{url} 
\usepackage[all]{hypcap} 

\usepackage[misc]{ifsym} 

\usepackage[noadjust]{cite}  
\newcommand{\keywords}[1]{%
  \vspace{10pt} \noindent \textbf{Keywords:} #1
}

\usepackage{xspace}
\newcommand{\mk}[1]{\textcolor{magenta}{\textbf{Murat:} #1}\xspace}
\renewcommand{\mk}[1]{}

\title{A Review of DeepSeek Models' Key Innovative Techniques}

\author{
    Chengen Wang \Envelope\\
    University of Texas at Dallas\\
    {\small\texttt{chengen.wang@utdallas.edu}}
    \and
    Murat Kantarcioglu\\
    Virginia Tech\\
    {\small\texttt{muratk@vt.edu}}
}

\date{} 

\begin{document}
\maketitle

\begin{abstract}
DeepSeek-V3 and DeepSeek-R1 are leading open-source Large Language Models (LLMs) for general-purpose tasks and reasoning, achieving performance comparable to state-of-the-art closed-source models from companies like OpenAI and Anthropic---while requiring only a fraction of their training costs. Understanding the key innovative techniques behind DeepSeek's success is crucial for advancing LLM research. In this paper, we review the core techniques driving the remarkable effectiveness and efficiency of these models, including refinements to the transformer architecture, innovations such as Multi-Head Latent Attention and Mixture of Experts, Multi-Token Prediction, the co-design of algorithms, frameworks, and hardware, the Group Relative Policy Optimization algorithm, post-training with pure reinforcement learning and iterative training alternating between supervised fine-tuning and reinforcement learning. Additionally, we identify several open questions and highlight potential research opportunities in this rapidly advancing field.
\end{abstract}


\section{Introduction}
The emergence of ChatGPT in late 2022~\cite{chatgpt} ushered in a new era of Large Language Model (LLM) research. LLMs have since advanced rapidly, with models like GPT~\cite{gpt_models} and Claude~\cite{claude} demonstrating exceptional performance. While open-source LLMs such as LLaMA~\cite{grattafiori2024llama} have achieved competitive results in certain metrics, their overall performance still lags behind proprietary models.

In January 2025, DeepSeek rattled markets and made headlines~\cite{stockmarket} with DeepSeek-V3~\cite{liu2024deepseek_v3} and newly-launched DeepSeek-R1 models~\cite{guo2025deepseek_R1}. These models achieve performance comparable to that of state-of-the-art GPT models, while requiring only a fraction of training resources. Understanding the techniques underlying the remarkable effectiveness and efficiency of these models is crucial for advancing LLM research.

In this paper, we review the key techniques behind DeepSeek Models' success. These include the refinement to the transformer architecture---specifically, Multi-Head Latent Attention (MLA) and Mixture of Experts (MoE); Multi-Token Prediction; the co-design of algorithms, frameworks and hardware; the Group Relative Policy Optimization (GRPO) reinforcement learning algorithm; and post-training techniques, such as pure reinforcement learning and multi-stage iterative training that alternates between Supervised Fine-Tuning (SFT) and reinforcement learning.

Additionally, we identify several issues that are not addressed in DeepSeek's technical report or ablation studies, highlighting potential research opportunities.

In the following, we first provide a concise yet in-depth review of the above-mentioned innovative techniques in Section~\ref{sec:tech}, followed by a discussion of the open problems and potential research directions in Section~\ref{sec:discuss}, and conclude the paper in Section~\ref{sec:conclude}.

\section{The Innovative Techniques} \label{sec:tech}
In this section, we examine the key innovative techniques that drive the success of the DeepSeek models. While these techniques are integrated into DeepSeek-V3 and DeepSeek-R1, some may have been introduced in earlier DeepSeek models.

\subsection{Multi-Head Latent Attention} \label{sec:mla}

KV cache is a technique used in the Multi-Head Attention (MHA) block of a transformer to accelerate inference by storing intermediate keys and values, eliminating the need for repeated calculations. However, the KV cache can become a bottleneck for long-context LLMs due to their high memory consumption. One approach to reducing the KV cache is to employ fewer attention heads, as seen in Multi-Query Attention (MQA)~\cite{shazeer2019fast_MQA} and Group-Query Attention (GQA)~\cite{ainslie2023gqa}. Despite this, their performance does not match that of MHA. Later, an innovative attention mechanism called Multi-head Latent Attention (MLA) is proposed for DeepSeek-V2~\cite{liu2024deepseek_v2}, which requires far less KV cache while achieving better performance. 

\subsubsection{Standard Multi-Head Attention}
In the standard MHA~\cite{vaswani2017attention}, the queries, keys and values are obtained through the projection matrices $W^Q, W^K, W^V\in\mathbb{R}^{d_h n_h\times d}$, transforming $\mathbf{h_t}\in\mathbb{R}^d$, the input of $t$-th token, to queries, keys and values $\mathbf{q}_t=W^Q\mathbf{h}_t,\mathbf{k}_t=W^K\mathbf{h}_t, \mathbf{v}_t=W^V\mathbf{h}_t, \mathbf{q}_t, \mathbf{k}_t, \mathbf{v}_t\in\mathbb{R}^{d_h n_h}$, respectively, where $d$ is the dimension of the input embedding, $n_h$ is the number of heads and $d_h$ is the dimension per head.

The dimension $d_h\times n_h$ indicates how the $\mathbf{q}_{t}, \mathbf{k}_{t}, \mathbf{v}_{t}$ are sliced into $n_h$ heads with dimension $d_h$ per head for the multi-head attention mechanism~\cite[Eq. (4)-(8)]{liu2024deepseek_v2}:
\begin{align}
    [\mathbf{q}_{t, 1};&\mathbf{q}_{t, 2};...;\mathbf{q}_{t, n_{h}}] = \mathbf{q}_{t}, \\
    [\mathbf{k}_{t, 1};&\mathbf{k}_{t, 2};...;\mathbf{k}_{t, n_{h}}] = \mathbf{k}_{t}, \\
    [\mathbf{v}_{t, 1};&\mathbf{v}_{t, 2};...;\mathbf{v}_{t, n_{h}}] = \mathbf{v}_{t}, \\
    \mathbf{o}_{t, i} &= \sum_{j=1}^{t} \operatorname{Softmax}_j(\frac{\mathbf{q}_{t, i}^T \mathbf{k}_{j, i}}{\sqrt{d_{h}}}) \mathbf{v}_{j, i}, \\ 
    \mathbf{u}_{t} &= W^{O} [\mathbf{o}_{t, 1};\mathbf{o}_{t, 2};...;\mathbf{o}_{t, n_{h}}],
\end{align}
where $\mathbf{q}_{t, i}, \mathbf{k}_{t, i}, \mathbf{v}_{t, i} \in \mathbb{R}^{d_h}$ represent the query, key, and value of the $i$-th head, respectively, and 
$W^{O} \in \mathbb{R}^{d \times d_h n_h}$ is the output projection matrix. 
During inference, each token requires KV cache of size $2 n_{h} d_{h} l$, where $l$ is the number of layers.

\begin{figure}[h] 
     \centering
     \includegraphics[width=0.9\textwidth]{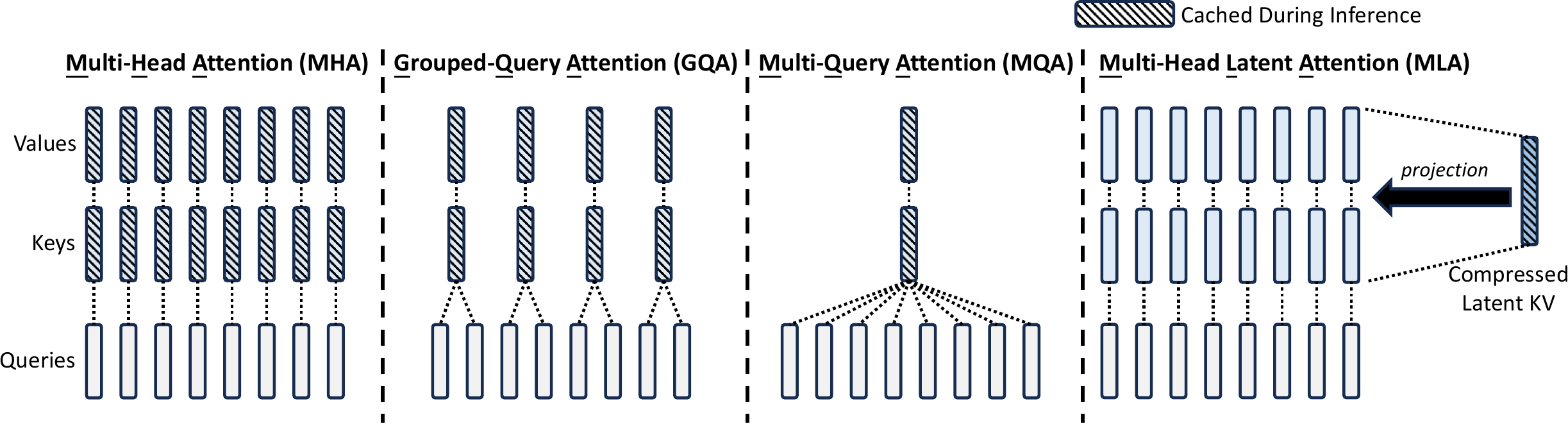}
     \caption{A simplified illustration of Multi-Head Attention (MHA), Grouped-Query Attention (GQA), Multi-Query Attention (MQA), and Multi-head Latent Attention (MLA). Adapted from~\cite[\textsc{Figure 3}]{liu2024deepseek_v2}.}
     \label{fig:mla}
\end{figure}
\subsubsection{Low-Rank Key-Value Joint Compression}
 The core idea of MLA is to decompose the projection matrix into two lower-rank matrices: $W=W^UW^{DKV}$, where $W^{DKV}\in\mathbb{R}^{d_c\times d}$ is the down-projection matrix for both keys and values, $W^U\in\mathbb{R}^{d_h n_h\times d_c}$ is the up-projection matrix, and $d_c\ll d_h n_h$. The down-projection matrix compresses \textit{both} keys and values into \textit{one} latent vector  $\mathbf{c}_t^{KV}=W^{DKV}\mathbf{h}_t, \mathbf{c}_t^{KV}\in\mathbb{R}^{d_c}$~\cite[Eq. (9)]{liu2024deepseek_v2}. Since $d_c\ll d_hn_h$, for each token, saving $\mathbf{c}_t^{KV}$, of size $d_c l$, instead of both $\mathbf{k}_t$ and $\mathbf{v}_t$, of size $2d_h n_h l$, greatly reduces the KV cache.
 
The keys and values are computed from the latent vector $\mathbf{c}_t^{KV}$ as follows~\cite[Eq. (10)-(11)]{liu2024deepseek_v2}:
\begin{align}
    \label{eq:c_to_k}
    \mathbf{k}_{t}^{C} &= W^{UK} \mathbf{c}_{t}^{KV}, \\
    \mathbf{v}_{t}^{C} &= W^{UV} \mathbf{c}_{t}^{KV}, 
\end{align}
where $W^{UK},W^{UV} \in \mathbb{R}^{d_h n_h \times d_c}$ denote the up-projection matrices for keys and values, respectively.
Importantly, $W^{UK}$ will be absorbed into $W^Q$ and $W^{UV}$ absorbed into $W^O$ during inference, so we do not need compute $\mathbf{k}_{t}^{C}, \mathbf{v}_{t}^{C}$ explicitly. The architecture of MLA is illustrated in Figure~\ref{fig:mla}.

Moreover, \textit{low-rank} compression of queries is applied to reduce the activation memory during training~\cite[Eq. (12)-(13)]{liu2024deepseek_v2}:
\begin{align}
    \mathbf{c}_{t}^{Q} &= W^{DQ} \mathbf{h}_{t}, \\
    \mathbf{q}_{t}^{C} &= W^{UQ} \mathbf{c}_{t}^{Q},
\end{align}
where $\mathbf{c}_{t}^{Q} \in \mathbb{R}^{d_c^{\prime}}$ represents the compressed latent vectors for queries, with $d_c^{\prime} \ll d_h n_h$, 
and $W^{DQ} \in \mathbb{R}^{d_c^{\prime} \times d}, W^{UQ} \in \mathbb{R}^{d_h n_h \times d_c^{\prime}}$ denote the down-projection and up-projection matrices, respectively. 

\subsubsection{Decoupled Rotary Position Embedding} \label{sec:rotary}
DeepSeek-V2 utilizes the Rotary Position Embedding (RoPE)~\cite{su2024roformer}:
\begin{align}
    \mathbf{q}_i^T\mathbf{k}_j&=\mathbf{h}_i^T(W^{Q})^T\operatorname{RoPE}_{\Theta,j-i}(W^{K}\mathbf{h}_j) \\
    &=\mathbf{h}_i^T(W^{Q})^T\operatorname{RoPE}_{\Theta,j-i}(W^{UK}W^{DKV}\mathbf{h}_j)
\end{align}
where $\operatorname{RoPE_{\Theta,j-i}(\cdot)}$ denotes the operation that applies the RoPE matrix, $\Theta$ is pre-defined parameters, and $i,j$ are the $i$-th and $j$-th positions.
As a result, $W^{UK}$ will not be absorbed into $W^{Q}$, leading to significant computational cost during inference.

To address this issue, DeepSeek-V2 proposes to decouple RoPE into a separate set of queries and keys: multi-head queries $\mathbf{q}_{t, i}^{R} \in \mathbb{R}^{d_h^R}$ and a key $\mathbf{k}_{t}^{R} \in \mathbb{R}^{d_h^R}$ \textit{shared} by all heads, where $d_h^R$ represents the per-head dimension of the decoupled queries and keys. This decoupling strategy essentially computes two separate sets of attention weights, which are then added together. The full MLA computation is as follows~\cite[Eq. (14)-(19)]{liu2024deepseek_v2}:
\begin{align}
    [\mathbf{q}_{t, 1}^{R};\mathbf{q}_{t, 2}^{R};...;\mathbf{q}_{t, n_{h}}^{R}] = \mathbf{q}_{t}^{R} &= \operatorname{RoPE}({W^{QR}} \mathbf{c}_{t}^{Q}), \\
    \mathbf{k}_{t}^{R} &= \operatorname{RoPE}({W^{KR}} \mathbf{h}_{t}), \\
    \mathbf{q}_{t, i} &= [\mathbf{q}_{t, i}^{C}; \mathbf{q}_{t, i}^{R}], \\
    \mathbf{k}_{t, i} &= [\mathbf{k}_{t, i}^{C}; \mathbf{k}_{t}^{R}], \\
    \mathbf{o}_{t, i} &= \sum_{j=1}^{t} \operatorname{Softmax}_j(\frac{\mathbf{q}_{t, i}^T \mathbf{k}_{j, i}}{\sqrt{d_{h} + d_{h}^{R}}}) \mathbf{v}_{j, i}^{C}, \\ 
    \mathbf{u}_{t} &= W^{O} [\mathbf{o}_{t, 1};\mathbf{o}_{t, 2};...;\mathbf{o}_{t, n_{h}}],
\end{align}
where $W^{QR} \in \mathbb{R}^{d_h^R n_h \times d_c^{\prime}}$ and $W^{KR} \in \mathbb{R}^{d_h^R \times d}$ denote matrices used to generate the decoupled queries and key, respectively, 
$\operatorname{RoPE}(\cdot)$ refers to the operation that applies the RoPE matrix, with the subscripts omitted,
and $[\cdot;\cdot]$ represents the concatenation operation.
During inference, the decoupled key $\mathbf{k}_{t}^{R}$ with dimension $d_{h}^{R}$ is also cached. As a result, each token requires cache of size $(d_c+d_{h}^{R})l$ in total. For DeepSeek-V2, where $d_c=4d_h$ and $d_h^R=\frac{d_h}{2}$, the KV cache per token is $\frac{9}{2}d_{h}l$.

It has been reported that MLA outperforms MHA~\cite[Table 9]{liu2024deepseek_v2}, which is surprising considering that MLA uses low-rank matrices, inherently containing less information than the original projection matrices for keys and values. Therefore, this performance gain is likely due to the introduction of the decoupled RoPE, which differs from the original RoPE. However, no ablation study for the decoupled RoPE has been reported, making it a worthwhile direction for further investigation.

\begin{figure}[h] 
     \centering
     \includegraphics[width=0.9\textwidth]{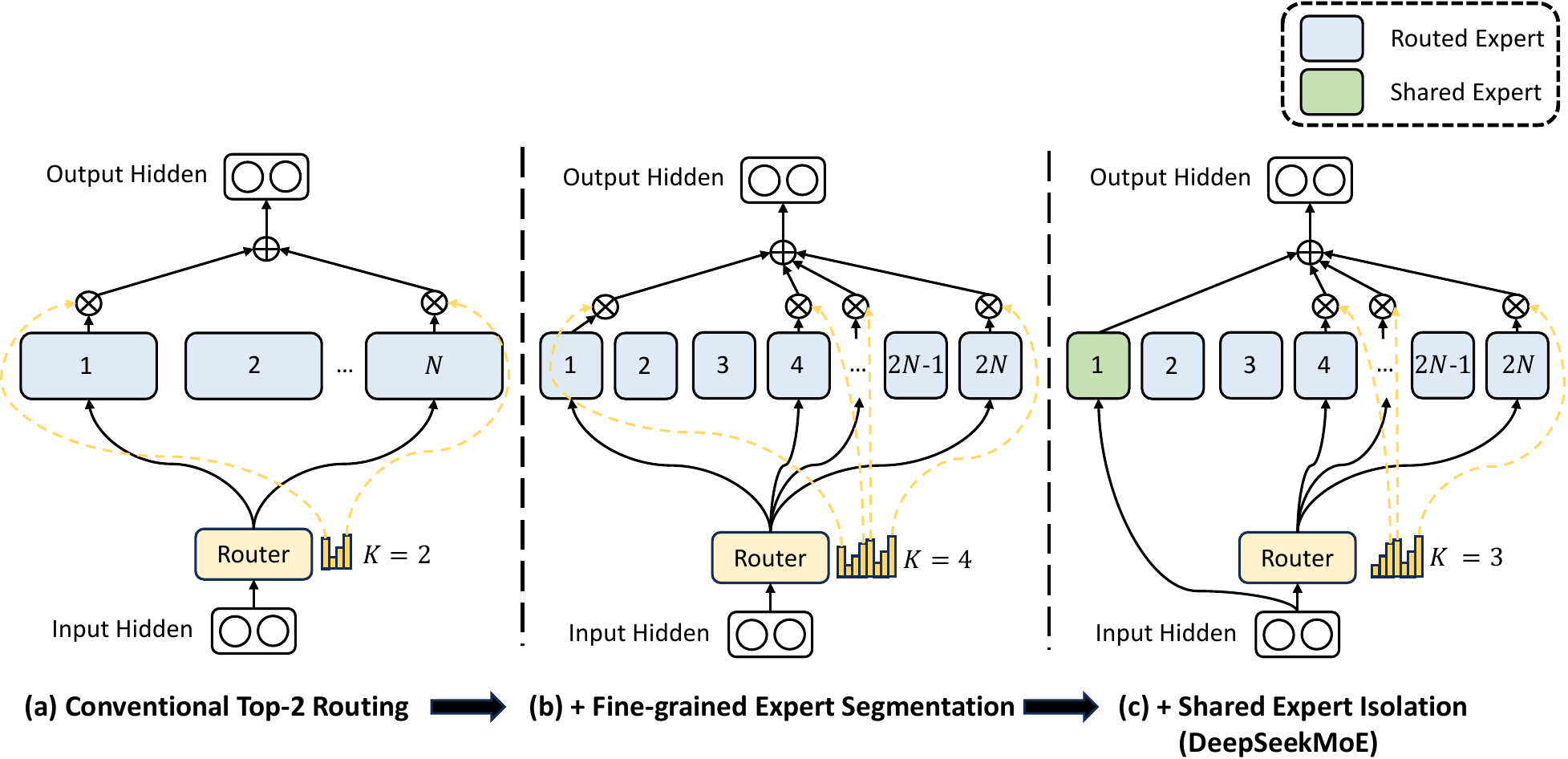}
     \caption{An illustration of DeepSeekMoE. Note the number of expert parameters and computational cost remain the same across the three architectures. Adapted from~\cite[\textsc{Figure 2}]{dai2024deepseekmoe}.}
     \label{fig:moe}
\end{figure}
\subsection{Mixture of Experts}
Mixture of Experts (MoE) is an architecture designed to reduce computational cost while scaling up model parameters. In an MoE model, the Feed-Forward Network (FFN) layers in a Transformer are typically replaced with MoE layers at specified intervals. Each MoE layer consists of multiple experts, all structurally identical to a standard FFN. Tokens are routed to one or two experts~\cite{fedus2022switch, lepikhin2020gshard}. The DeepSeekMoE architecture~\cite{dai2024deepseekmoe} introduces two key innovations: fine-grained expert segmentation and shared expert isolation. These innovations are built upon the conventional MoE.

\subsubsection{Fine-Grained Expert Segmentation}
On top of the conventional MoE architecture shown in Figure~\ref{fig:moe}(a), each FFN is segmented into $m$ smaller experts by dividing the FFN hidden dimension evenly. As a result, if the total number of experts is $N$ and the number of activated experts for each token is $K$ in a conventional MoE, then the total number of experts is increased to $mN$ and the number of activated experts is increased to $mK$ for a fine-grained MoE architecture, as illustrated in Figure~\ref{fig:moe}(b). This fine-grained segmentation strategy greatly improves the combinatorial flexibility of the activated experts. 

\subsubsection{Shared Expert Isolation}
Shared experts are dedicated to capture the common knowledge across diverse contexts, reducing parameter redundancy among different experts. Specifically, $K_s$ experts are reserved as shared experts, and each token will be always assigned to these shared experts in additional to their respective routed experts. To maintain a constant computational cost, the $total$ number of \textit{routed} experts $N_r$ is reduced to $mN-K_s$ and the number of routed experts for each token is $mK-K_s$.

With the novel strategy of fine-grained expert segmentation and shared expert isolation, an MoE layer in the DeepSeekMoE architecture is defined as follows~\cite[Eq. (9)-(11)]{dai2024deepseekmoe}:
\begin{align}
\mathbf{h}_{t}^{l} & = \sum_{i=1}^{K_{s}} {\operatorname{FFN}_{i}\left( \mathbf{u}_{t}^{l} \right)} + \sum_{i=K_{s} + 1}^{mN} \left( {g_{i,t} \operatorname{FFN}_{i}\left( \mathbf{u}_{t}^{l} \right)} \right) + \mathbf{u}_{t}^{l}, \\
g_{i,t} & = \begin{cases} 
s_{i,t}, & s_{i,t} \in \operatorname{Topk} (\{ s_{j, t} | K_{s} + 1 \leq j \leq mN \}, mK - K_{s}), \\
0, & \text{otherwise}, 
\end{cases} \\
s_{i,t} & = \operatorname{Softmax}_i \left( {\mathbf{u}_{t}^{l}}^{T} \mathbf{e}_{i}^{l} \right), 
\end{align}
where $\operatorname{FFN}_i(\cdot)$ refers to the $i$-th expert FFN, 
$\mathbf{u}_{t}^{l} \in \mathbb{R}^{d}$ is the hidden state of the $t$-th token after the $l$-th attention module, 
and $\mathbf{h}_{t}^{l} \in \mathbb{R}^{d}$ is the output hidden state of the $t$-th token after the $l$-th MoE layer.
$g_{i,t}$ represents the gate value for the $i$-th expert, 
$s_{i,t}$ is the token-to-expert affinity, 
$\operatorname{Topk}(\cdot, K)$ gives the set of top $K$ affinity scores calculated for the $t$-th token across all $N$ experts,
and $\mathbf{e}_{i}^{l}$ represents the centroid of the $i$-th expert in the $l$-th layer. 

\subsubsection{Load Balancing} \label{sec:balance}
The automatically learned routing strategy may face the issue of load imbalance, where either a few experts are always selected while others are not sufficiently trained, or the activated experts are distributed across multiple devices, leading to significant inter-device communication cost. These issues are address by an auxiliary loss for load balancing~\cite{fedus2022switch}. The expert-level balance loss is formulated as follows~\cite[Eq. (12)-(14)]{dai2024deepseekmoe}:
\begin{align}
\mathcal{L}_{\mathrm{ExpBal}} & = \alpha \sum_{i=1}^{N^{\prime}}{f_i P_i}, \\
    f_i & = \frac{N^{\prime}}{K^{\prime}T} \sum_{t=1}^{T}{ \mathds{1}( \text{Token $t$ selects Expert $i$} )}, \\
    P_i & = \frac{1}{T} \sum_{t=1}^{T}{s_{i,t}},
\end{align}
where $\alpha$ is a hyper-parameter, 
$N^{\prime}=mN - K_s$ and $K^{\prime}=mK - K_s$ for simplicity,
and $\mathds{1}(\cdot)$ represents the indicator function. When the load is uniformly-distributed among the experts, $\mathcal{L}_{\mathrm{ExpBal}}$ is minimized, $f_i=1, P_i=\frac{K'}{N'}$, and we have $\sum_{i=1}^{N^{\prime}}{f_i P_i}=N'\cdot 1\cdot \frac{K'}{N'}=K'$.

Let $f'_i$ and $P'_i$ be the normalized version of $f_i$ and $P_i$, respectively, $f'_i=\frac{f_i}{N'}, P'_i=\frac{P_i}{K'}$, such that both form probability distributions. The limitation of this expert-level loss formulation is that when $P'_i$ is uniformly-distributed, i.e., $P'_i=\frac{1}{N'}$ for all $i$, then for \textit{any} distribution of $f'_i$, we have $\sum_{i=1}^{N^{\prime}}{f_i\cdot P_i}=\sum_{i=1}^{N'}{f'_i}\cdot K'=K'$. Under these circumstances, the auxiliary loss fails to push toward balanced experts utilization. If a uniform distribution of $P'_i$ inherently produce a uniform distribution of $f'_i$ in practice, then incorporating $f_i$ in the loss function appears redundant. Given the widespread adoption of this formulation~\cite{fedus2022switch, liu2024deepseek_v3, jin2024moe++}, it is worthwhile to investigate its theoretical justification and explore potential improvements.

Besides expert-level load balancing, device-level and communication load balancing~\cite{dai2024deepseekmoe, liu2024deepseek_v2} are proposed to ensure balanced computation and communication across different devices. The formulations of these loss functions follow a similar pattern.

Since the auxiliary loss may degrade model performance, an auxiliary-loss-free load balancing strategy is proposed to strike a better trade-off between load balance and model performance~\cite{wang2024auxiliary-loss-free}. Specifically, a bias term $b_i$ for each expert $i$ is added to the affinity score $s_{i,t}$ to determine the top-K selection~\cite[Eq. (16)]{liu2024deepseek_v3}:
\begin{align}
    g^{\prime}_{i,t} & = \begin{cases} \label{eq:loss-free} 
    s_{i,t}, & s_{i,t} + b_i \in \operatorname{Topk} (\{ s_{j, t} + b_j | 1 \leq j \leq N_r \}, K_{r}), \\
    0, & \text{otherwise},
    \end{cases}
\end{align}
where $N_r$ represents the number of routed experts and $K_r$ is the number of activated routed experts. 
During training, the bias term $b_i$ will be decreased by $\gamma$ if the expert is overloaded and increased by $\gamma$ if the expert is underloaded, where $\gamma$ is a hyperparameter. Note that the bias term is used solely for top-K selection and the gating value is still using the original affinity score $s_{i,t}$, as shown in Eq.~(\ref{eq:loss-free}). In this equation, $ s_{i,t} + b_i $ serves as the input to the Topk($\cdot$) function, while $s_{i,t}$ is the value of $g_{i,t}^{\prime}$ if $s_{i,t}$ is among the top-K. In DeepSeek-V3, a complementary sequence-wise auxiliary loss is also employed to avoid extreme imbalance within any single sequence~\cite{liu2024deepseek_v3}.

\begin{figure}[h] 
     \centering
     \includegraphics[width=0.9\textwidth]{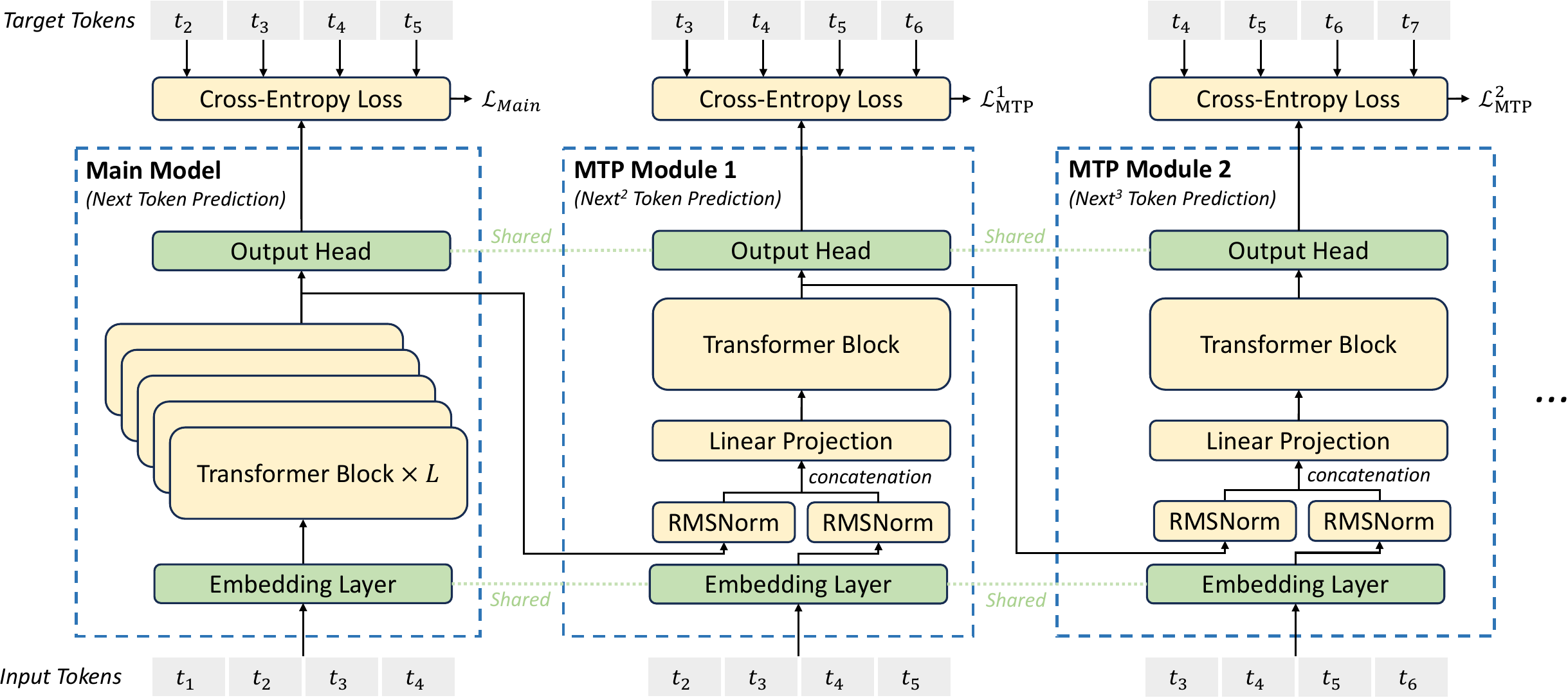}
     \caption{An illustration of the Multi-Token Prediction (MTP) implementation in DeepSeek-V3, which keeps the
complete causal chain for predicting each token at every depth. Adapted from~\cite[\textsc{Figure 3}]{liu2024deepseek_v3}.}
     \label{fig:mtp}
\end{figure}
\subsection{Multi-Token Prediction} \label{sec:multi-token}
DeepSeek-V3 employs Multi-Token Prediction (MTP)~\cite{gloeckle2024better_multi_token} to improve training performance. For each token, instead of predicting next-token only, MTP predicts $D$ additional tokens in a causal chain, as shown in Figure~\ref{fig:mtp}. At each depth $k$ of the $D$ MTP modules, there are a shared Embedding Layer and a shared Output Head, an independent Transformer Block, and an independent Linear Projection layer. The input to the Linear Projection layer is an concatenation of the embedding at the current depth and the output embedding from the previous depth.

The MTP training objective, $\mathcal{L}_{\text{MTP}}$, is the average of the cross-entropy loss $\mathcal{L}_{\text{MTP}}^{k}$ at each depth $k\in\{1,2,\cdots,D\}$~\cite[Eq. (24)-(25)]{liu2024deepseek_v3}:
\begin{align}
    \mathcal{L}_{\text{MTP}}^{k} = \operatorname{CrossEntropy}(P_{2 + k:T + 1}^{k}, t_{2 + k:T + 1}) = -\frac{1}{T} \sum_{i=2 + k}^{T + 1} \log P_i^k [t_i],\\
    \mathcal{L}_{\text{MTP}} = \frac{\lambda}{D} \sum_{k=1}^{D} \mathcal{L}_{\text{MTP}}^{k},
\end{align}
where $T$ represents the length of the input sequence, $t_i$ is the ground-truth token at the $i$-th position, and $P_i^k [t_i]$ denotes the prediction probability of $t_i$ at depth $k$.

The advantage of MTP lies in its higher sample efficiency during training~\cite{gloeckle2024better_multi_token}, leading to improved performance. However, the causal chain formed by the MTP modules introduces additional \textit{training} time overhead beyond conventional next-token prediction, a factor not addressed in the ablation study for MTP in DeepSeek-V3~\cite[Sec. 4.5.1]{liu2024deepseek_v3}.

\subsection{Co-design of Algorithms, Frameworks and Hardware}
Through the co-design of algorithms, frameworks and hardware, with meticulous engineering optimizations, DeepSeek-V3 significantly enhances the training efficiency and completes the pre-training of the model on $14.8$ trillion tokens with $2.788$ million H800 GPU hours~\cite{liu2024deepseek_v3}.

\begin{figure}[h] 
     \centering
     \includegraphics[width=0.9\textwidth]{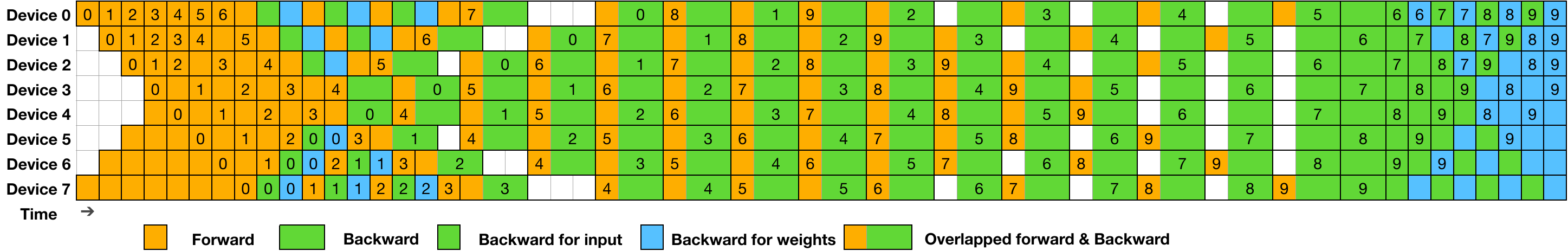}
     \caption{An example of DualPipe scheduling for 8 PP ranks and 20 micro-batches in two directions. The micro-batches in the reverse direction are symmetric to those in the forward direction. Adapted from~\cite[\textsc{Figure 5}]{liu2024deepseek_v3}.}
     \label{fig:dualpipe}
\end{figure}
\subsubsection{DualPipe} \label{sec:dualpipe}
To reduce the communication overhead introduced by cross-node expert parallelism, an innovative pipeline parallelism algorithm called DualPipe~\cite{liu2024deepseek_v3} is proposed to overlap the computation and communication within a pair of individual forward and backward chunks. The algorithm divides each chunk into four components, with the backward computation chunk further divided into two parts for input and weights, respectively~\cite{qi2023zero}, to reduce pipeline bubbles. A specific ratio of GPU SMs are dedicated to communication, ensuring that communication remains fully hidden during execution, effectively achieving near-zero all-to-all communication overhead. The DualPipe algorithm employs a bidirectional pipeline scheduling, feeding data from both ends of the pipeline, as illustrated in Figure~\ref{fig:dualpipe}.

DualPipe requires keeping two copies of the models parameters, leading to additional memory consumption. It turns out the \textit{bidirectional} part is unnecessary and can be removed with a ``cut-in-half" procedure, as outlined in~\cite{qi2025dual_improved}.


\subsubsection{FP8 Mixed Precision Training}
A mixed precision framework for training DeepSeek-V3 is introduced for efficient training without accuracy degradation. In order to accelerate training, the \textit{majority} of core computation kernels---General Matrix Multiplication (GEMM)---are implemented in FP8 precision~\cite{dettmers2022gpt3_int8, peng2023fp8, fishman2024scaling_fp8}.
Despite the efficiency advantage of FP8 format, DeepSeek-V3 maintains the original precision for certain operators due to their sensitivity to low-precision computations, to balance training efficiency and numerical stability. These operators include the embedding module, the output head, MoE gating modules, normalization operators, and attention operators.

This framework utilizes a fine-grained quantization strategy to extend the \textit{dynamic range} of the FP8 format: tile-wise grouping with $1\times N_c$ elements or block-wise grouping with $N_c\times N_c$ elements, where $N_c$ is the channel size, with $N_c=128$ in DeepSeek-V3 model.

The accuracy of low-precision GEMM operations largely depends on high-precision \textit{accumulation}. DeepSeek-V3 employs a strategy of promotion to CUDA Cores for higher precision, periodically copying intermediate results to FP32 registers on CUDA Cores at an interval $N_c$ for full-precision FP32 accumulation.

\begin{figure}[h] 
     \centering
     \includegraphics[width=0.9\textwidth]{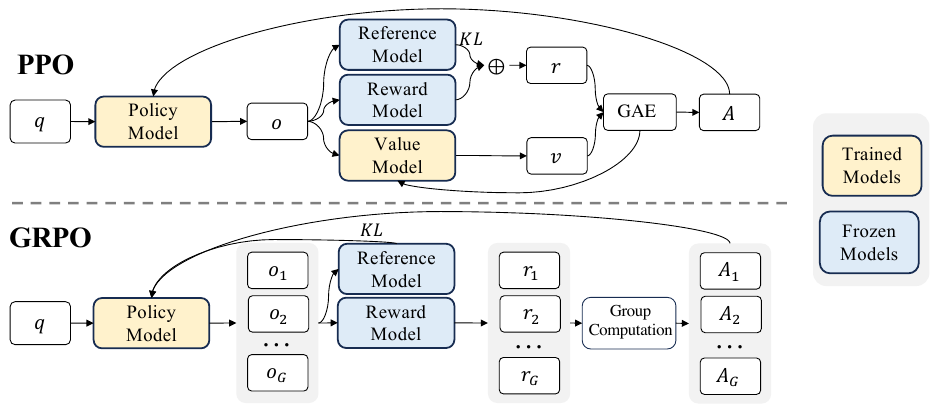}
     \caption{A comparison of PPO and GRPO. GRPO estimates the baseline from group scores, without a value model. Adapted from~\cite[\textsc{Figure 4}]{shao2024deepseekmath}.}
     \label{fig:grpo}
\end{figure}
\subsection{Group Relative Policy Optimization} \label{sec:grpo}

Group Relative Policy Optimization (GRPO)~\cite{shao2024deepseekmath} is an efficient and effective variant of Proximal Policy Optimization (PPO)~\cite{schulman2017proximal_ppo}. GRPO eliminates the value function approximation in PPO by directly estimating the advantage, significantly reducing memory usage. In the context of LLM, where typically only the last token is assigned a reward, the training of a value function in PPO is challenging. The simplified GRPO can achieve comparable performance while being more efficient.

More specifically, the PPO maximizes the following objective~\cite[Eq. (1)]{shao2024deepseekmath}:
\begin{equation}
\footnotesize
    \mathcal{J}_{PPO}(\theta) = \mathbb{E}{[q \sim P(Q), o \sim \pi_{\theta_{old}}(O|q)]} \frac{1}{|o|} \sum_{t=1}^{|o|} \min \left[ \frac{\pi_\theta(o_{t} | q, o_{<t})}{\pi_{\theta_{old}}(o_{t} | q, o_{<t})} A_{t}, \text{clip} \left( \frac{\pi_\theta(o_{t} | q, o_{<t})}{\pi_{\theta_{old}}(o_{t} | q, o_{<t})}, 1 - \epsilon, 1 + \epsilon \right)  A_{t} \right] ,
\end{equation}
where $q, o$ represent questions and outputs, $\pi_{\theta}$ and $\pi_{\theta_{old}}$ are the current and old policy models, respectively, and $\epsilon$ is a hyper-parameter related to clipping, and $A_t$ represents the advantage, which is estimated using rewards and a learned value function.

Instead of training a value function, GRPO directly estimates the advantage by sampling a group of outputs $\{o_1, o_2, \cdots, o_G\}$ from the old policy $\pi_{\theta_{old}}$, yielding $G$ rewards $\mathbf{r}=\{r_1, r_2,\cdots,r_G\}$, scored by a reward model. There are two ways to estimate the advantage: outcome supervision and process supervision. Output supervision provides a reward at the end of each output $o_i$, and set the advantages $\hat{A}_{i,t}$ of all tokens to the same normalized reward, i.e., $\hat{A}_{i, t} = \widetilde{r}_i = \frac{r_i- {\rm mean}(\mathbf{r})}{{\rm std}(\mathbf{r})}$~\cite[Sec. 4.1.2]{shao2024deepseekmath}. Process supervision provides a reward for each intermediate steps, and then calculate the advantage for each token by summing the normalized rewards obtained from the subsequent steps. 

GRPO maximizes the following objective~\cite[Eq. (3)]{shao2024deepseekmath}:
\begin{equation}
\footnotesize
\begin{split}
    \mathcal{J}_{GRPO}(\theta) &= \mathbb{E}{[q \sim P(Q), \{o_i\}_{i=1}^G \sim \pi_{\theta_{old}}(O|q)]}  \\
    & \frac{1}{G}\sum_{i=1}^G\frac{1}{|o_i|} \sum_{t=1}^{|o_i|} \left\{ \min \left[ \frac{\pi_\theta(o_{i,t} | q, o_{i,<t})}{\pi_{\theta_{old}}(o_{i,t} | q, o_{i,<t})} \hat{A}_{i,t}, \text{clip} \left( \frac{\pi_\theta(o_{i,t} | q, o_{i,<t})}{\pi_{\theta_{old}}(o_{i,t} | q, o_{i,<t})}, 1 - \epsilon, 1 + \epsilon \right)  \hat{A}_{i,t} \right] - \beta \mathbb{D}_{KL}\left[\pi_{\theta} || \pi_{ref}\right]\right\},
\end{split}
\label{eq:GRPO-obj}
\end{equation}
where $\pi_{ref}$ is the reference policy model, typically the initial base model or the SFT model.
The GRPO algorithm is illustrated in Figure~\ref{fig:grpo}.

\subsection{Post-Training: Reinforcement Learning on the Base Model}
\subsubsection{Pure Reinforcement Learning}
DeepSeek-R1-Zero~\cite{guo2025deepseek_R1} is trained on the base model DeepSeek-V3-Base with pure reinforcement learning (RL) without any supervised fine-tuning (SFT) data. The performance of DeepSeek-R1-Zero continuously improves throughout the RL training process. The reasoning behaviors, including reflection and the exploration of alternative approaches, naturally arise during the training, demonstrating the effectiveness of pure RL and the model's capability to learn and generalize solely through RL.

DeepSeek-R1-Zero utilizes the GRPO algorithm, as outlined in Section~\ref{sec:grpo}. The reward function includes two types of rewards: \textit{accuracy rewards}, which assess the correctness of the model's response, and \textit{format rewards}, which enforces the model to enclose its thinking process within the tags ``\verb|<think>|" and ``\verb|</think>|". A training template is designed to guide the model in following a specified format, first generating a reasoning process before presenting the final answer.

Although DeepSeek-R1-Zero achieves remarkable performance through pure RL, it encounters challenges such as poor readability and language mixing. To mitigate these issues and further enhance the model, DeepSeek-R1~\cite{guo2025deepseek_R1} is introduced, which employs an iterative training approach that alternates between SFT and RL.

\subsubsection{Reinforcement Learning with Cold Start}
DeepSeek-R1 employs a training pipeline consisting of four stages~\cite{guo2025deepseek_R1}:
\begin{itemize}
    \item \textbf{Cold Start} To mitigate the instability of the early cold start phase of RL training, thousands of long Chain-of-Thought (CoT)~\cite{wei2022chain_CoT} examples are collected to fine-tune the DeepSeek-V3-Base model, which then serves as the foundation for subsequent reinforcement learning.
    \item \textbf{Reasoning-oriented RL} After fine-tuning DeepSeek-V3-Base on the cold-start data, the model undergoes the same RL training  process as employed in DeepSeek-R1-Zero. To address language mixing, an additional language consistency reward is introduced, measured as the proportion of target words in the CoT.
    \item \textbf{Rejection Sampling and SFT} The goal of this phase is to improve the model's performance in writing, role-playing and other general-purpose tasks. Once Reasoning-oriented RL converges, $600$k reasoning-related training samples are collected via rejection sampling from the checkpoint, retaining only the correct responses. Additionally, approximately $200$k non-reasoning training samples are collected, either from parts of the SFT dataset of DeepSeek-V3 or generated by DeepSeek-V3.
    \item \textbf{RL Alignment} This phase aims to better align the model with human preferences, by improving its helpfulness and harmlessness while also refining its reasoning capabilities. Helpfulness is measure based on the utility and relevance of the response, while harmlessness is evaluated throughout the entire response to reduce potential risks, biases or harmful content.
\end{itemize}



\section{Discussions} \label{sec:discuss}
In this section, we identify several areas where DeepSeek has made innovations and highlight potential future directions.
\begin{itemize}    
    \item \textbf{Transformer Architecture Improvement} The transformer serves as the core building block of LLMs. While MLA improves the attention mechanism, MoE enhances the FFN block within the transformer. Together, these innovations have significantly contributed to the advancement of the DeepSeek-V3 model. Advancements in transformer architecture can greatly influence both the effectiveness and efficiency of the training process. For example, a comprehensive ablation study of the decoupled rotary position embedding, as discussed in Section~\ref{sec:rotary}, could provide deeper insights. Additionally, further theoretical justification for the load balancing objective, as outlined in Section~\ref{sec:balance}, would be valuable for future research.
    \item \textbf{High Sample Efficiency} The introduction of multi-token prediction enhances the utilization of training data, thereby improving sample efficiency~\cite{gloeckle2024better_multi_token}. This demonstrates that better training efficiency can be achieved by developing algorithms that make more effective use of the training dataset. In Section~\ref{sec:multi-token}, we mention the issue of the incurred longer training time, suggesting there is still room for further improvement.
    \item \textbf{Co-design of algorithms, frameworks and hardware} The DualPipe and FP8 mixed precision training are engineering techniques introduced to enhance training efficiency. These innovations emphasize the value of designing models from a holistic perspective, integrating architecture, algorithms and hardware in the most effective and efficient manner. Recently, an improvement to the DualPipe has been made by~\cite{qi2025dual_improved}, as mentioned in Section~\ref{sec:dualpipe}.
    \item \textbf{Reinforcement Learning} The impressive performance of pure reinforcement learning in the post-training stage highlights a new research avenue in this area. The iterative training approach, which alternates between SFT and RL, is particularly inspiring. Furthermore, the introduction of GRPO demonstrates how a widely used RL algorithm can be improved to significantly reduce GPU memory usage.
\end{itemize}

\section{Conclusion} \label{sec:conclude}
In this paper, we have reviewed the key innovative techniques that contributed to the success of DeepSeek models. These include innovations in the transformer architecture, techniques for improving sample efficiency, the co-design of algorithms, frameworks and hardware, as well as the GRPO reinforcement learning algorithm and the application of reinforcement learning in the post-training stage. Our review highlights several open questions and potential research directions in this rapidly advancing field.

\addcontentsline{toc}{section}{References} 
\bibliographystyle{alpha}
\bibliography{references}

\end{document}